# Unbiased Statistics of a Constraint Satisfaction Problem – a Controlled-Bias Generator

Denis Berthier

Institut Telecom ; Telecom & Management SudParis

9 rue Charles Fourier, 91011 Evry Cedex, France

**Abstract:** We show that estimating the complexity (mean and distribution) of the instances of a fixed size Constraint Satisfaction Problem (CSP) can be very hard. We deal with the main two aspects of the problem: defining a measure of complexity and generating random unbiased instances. For the first problem, we rely on a general framework and a measure of complexity we presented at CISSE08. For the generation problem, we restrict our analysis to the Sudoku example and we provide a solution that also explains why it is so difficult.

## I. INTRODUCTION

Constraint Satisfaction Problems (CSP) constitute a very general class of problems. A finite CSP is defined by a finite number of variables with values in fixed finite domains and a finite set of constraints (i.e. of relations they must satisfy); it consists of finding a value for each of these variables, such that they globally satisfy all the constraints. General solving methods are known [1, 2]. Most of these methods combine a blind search algorithm (also called depth-first or breadth-first structured search, Trial and Error with Guessing, …) with some form of pattern-based pruning of the search graph.

In [3, 4, 5], we introduced a new general framework, based on the idea of a constructive, fully pattern-based solution and on the concepts of a candidate (a value not yet known to be impossible) and a resolution rule (which allows to progressively eliminate candidates). In [6], we introduced several additional notions, also valid for any CSP, such as those of a chain and a whip, and we showed how these patterns lead to general and powerful kinds of resolution rules.

The present paper relies on these general concepts (that are briefly recalled in order to make it as self-contained as possible) to analyse another question: *how can we define a measure of complexity for the instances of a given "fixed size" CSP and how can we estimate the statistical distribution of this complexity measure?* As yet, this question has received little interest and it could hardly have, because any method allowing blind search will rely on chance and hide the complexity of the various instances. With our constructive resolution approach, we can define a realistic mesure of complexity.

It should be clear that the above question is independent of a widely investigated problem, the NP-completeness of some types of CSPs. NP-completeness [7] supposes the CSP has a parameter (such as the size of a Sudoku grid: 9x9, 16x16; or the number of resources and tasks in a resource allocation problem) and one concentrates on worst case analysis as a function of this parameter. Here, on the contrary, we fix this parameter (if any), we consider the various instances of this fixed size CSP (e.g. all the 9x9 Sudoku puzzles) and we are more interested in mean case than in worst case analysis.

## II. MINIMAL INSTANCES

Instances of a fixed size CSP are defined by their givens (or clues): a *given* is a value pre-assigned to a variable of the CSP.

Instances of a CSP with several solutions cannot be solved in a purely constructive way: at some point, some choice must be made. Such under-constrained instances can be considered as ill-posed problems. We therefore concentrate on instances with a single solution.

It should also be obvious that, given an instance of a CSP, the more givens are added to it, the easier the resulting instances should become – the limit being when all the non given variables have only one possible value. This leads to the following definition: an instance of a CSP is called *minimal* if it has one and only one solution and it would have several solutions if any of its givens was deleted.

In statistical analyses, only samples of minimal instances are meaningful because adding extra givens would multiply the number of easy instances. We shall show that *building random unbiased samples of minimal instances may be very hard*.

## III. ZT-WHIPS AND THE ASSOCIATED MEASURE OF COMPLEXITY

The following definitions were introduced in [3], in the Sudoku context, and generalised to the general CSP in [6].

Definition: two different candidates of a CSP are *linked* by a direct contradiction (or simply linked) if one of the constraints of the CSP directly prevents them from being true at the same time *in any state* in which they are present (the fact that this notion does not depend on the state is fundamental). If two candidates are not linked, they are said *compatible*.

For any CSP, two different candidates for the same variable are always linked; but there may be additional direct contradictions; as expliciting them is part of modelling the CSP, we consider them as given with the CSP.

In Sudoku, two different candidates $n_1r_1c_1$ and $n_2r_2c_2$ are linked if: ($n_1 \neq n_2$ & $r_1c_1 = r_2c_2$) or ($n_1 = n_2$ & share-a-unit($r_1c_1$, $r_2c_2$)), where "share-a-unit" means "in the same row or in the same column or in the same block".

*A. zt-whips in a general CSP*

Definition: given a candidate Z (which will be called the target), a *zt-whip* of length n built on Z is a sequence $L_1$, $R_1$, $L_2$, $R_2$, … $L_n$, of 2n-1 (notice that there is no $R_n$) *different* candidates (alternatively called left-linking and right-linking candidates) for possibly different variables, such that, additionally:

for any $1 \leq k \leq n$, $L_k$ is linked to $R_{k-1}$ (setting $R_0 = Z$),

for any $1 \leq k < n$, $L_k$ and $R_k$ are candidates for the same variable (and they are therefore linked),

$R_k$ is the only candidate for this variable compatible with Z and with the previous right-linking candidates (i.e. with all the $R_j$, for j < k), for the same variable as $L_n$, there is no candidate compatible with the target and the previous right-linking candidates.

***zt-whip theorem for a general CSP: in any knowledge state of any CSP, if Z is a target of a zt- whip of any length, then it can be eliminated (formally, this rule concludes ¬Z).*** The proof was given in [6].

*B. The ZT measure of complexity*

For any CSP, we are now in a position to define an increasing sequence of theories (i.e. sets of resolution rules) based on zt-whips, an increasing sequence of sets of minimal puzzles solved by these theories and a rating for these instances:

$L_0$ is the set of resolution rules expressing the propagation of constraints (elimination of candidates due to the presence of a value for a variable) and of resolution rules asserting values for variables that have a unique candidate left;

for any n>0, $L_n$ is the union of $L_0$ with set of resolution rules for whips of length ≤ n.

as there can be no confusion between sets of rules and sets of instances, $L_n$ is also used to name the set of minimal instances of the CSP that can be solved with rules in $L_n$;

given an instance of a CSP, its *ZT rating* is defined as the smallest n such that this instance is in $L_n$.

In Sudoku, the zt-rating has a nice structural property: it is invariant under the (n, r, c) natural super-symmetries of the game, i.e two puzzles that are isomorphic under any of these symmetries have the same zt-rating. For this reason, we named zt-whips nrczt-whips [4, 5] and the zt-rating NRCZT.

There was an anterior measure of complexity, the SER, based on a very different approach and compatible with the players intuition of complexity, but not invariant under symmetries. It appears that the correlation coefficient (computed on several collections of a million puzzles each) between the NRCZT and the SER is always high: 0.895.

Finally, there is also a very good correlation between the NRCZT and the logarithm of the number of partial whips used in the resolution process: 0.946. This number is an intuitive measure of complexity, because it indicates among how many useless whips the useful ones must be found.

These two properties show that the NRCZT rating is a good (logarithmic) measure of complexity, from both theoretical and pragmatic points of view. We can therefore conclude that the first task we had set forth is accomplished.

*C. First statistical results for the Sudoku nrczt-whips*

In the Sudoku case, we have programmed all the rules for nrczt-whips in our SudoRules solver, a knowledge based system, running indifferently on the CLIPS [8] or the JESS [9] inference engine.

The following statistics are relative to a sample of one million puzzles obtained with the suexg [10] top-down random generator. This was our first, naive approach to the generation problem: using a generator of random minimal puzzles widely available and used by the Sudoku community.

Row 2 of Table 1 below gives the number of puzzles with NRCZT rating n. Row 3 gives the total number of puzzles solved when whips of length ≤ n (corresponding to resolution theory $L_n$) are allowed. This shows that more than 99% of the puzzles can be solved with whips of length ≤ 5 and more than 99.9% with whips of length ≤ 7. But there remain a few exceptional cases with much larger complexity.

| 0 | 1 | 2 | 3 | 4 | 5 | 6 | 7 | 8 | 9 | 10 | 11 | 12 | 13 |
|---|---|---|---|---|---|---|---|---|---|---|---|---|---|
| 417,624 | 120,618 | 138,371 | 168,562 | 122,946 | 24,187 | 5,511 | 1,514 | 473 | 130 | 38 | 15 | 9 | 2 |
| 417,624 | 538,242 | 676,613 | 845,175 | 968,121 | 992,308 | 997,819 | 999,333 | 999,806 | 999,936 | 999,974 | 999,989 | 999,998 | 1,000,000 |

Table 1: Number of puzzles in 1,000,000 with NRCZT rating n (row2) and solved with nrczt-whips of length ≤ n (row 3).

## IV. STANDARD TOP-DOWN AND BOTTOM-UP GENERATORS

A little after the above results were obtained, additional statistics led to suspect that the top-down suexg generator may have some bias. There is a very simple procedure for generating an unbiased Sudoku puzzle:
```
1) generate a random complete grid P;
2) for each cell in P, delete its value with
probability ½, thus obtaining a puzzle Q;
3)if Q is minimal, return it, otherwise goto 1.
```
Unfortunately, the probability of getting a valid puzzle this way is infinitesimal and one has to rely on other generators. Before going further, we introduce the two classical algorithms for generating minimal puzzles: bottom-up and top-down.

### A. *The classical bottom-up and top-down generators [12]*

A standard bottom-up generator works as follows to produce *one* minimal puzzle (it has to be iterated n times to produce n minimal puzzles):
```
1) start from an empty grid P
2a) in P, randomly choose an undecided cell and a
value for it, thus getting a puzzle Q;
2b) if Q is minimal, return it and exit;
2b) if Q has several solutions, set P = Q and GOTO 2a;
2c) if Q has no solutions, then goto 2a (i.e.
backtrack, forget Q and try another cell).
```

A standard top-down generator works as follows to produce *one* minimal puzzle (it has to be iterated n times to produce n minimal puzzles):
```
1) choose randomly a complete grid P;
2a) choose one clue randomly from P and delete it,
thus obtaining a puzzle P2;
2b) if P2 has several solutions, GOTO 2a (i.e.
reinsert the clue just deleted and try deleting
another);
2c) if P2 is minimal, printout P2 and exit the whole
procedure;
2d) otherwise (the puzzle has more than one solution),
set P=P2 and GOTO 2a.
```

Clause 2c in the bottom-up case and clause 2b in the top-down case make any analysis very difficlut. Moroever, it seems that they also cause the generator to look for puzzles with fewer clues. It may thus be suspected of introducing a strong, uncontrolled bias with respect to the number of clues.

### C. *Existence of a bias and a (weak) correlation*

The existence of a (as yet non measurable) bias in the number-of-clues distribution may in itself introduce a bias in the distribution of complexities (measured by the NRCZT or SER ratings). This bias may not be very large, as the correlation coefficient between the number of clues and the NRCZT or the SER was estimated (on our 1,000,000-puzzle sample) to be only 0.12. But it cannot be completely neglected either because it is an indication that other kinds of bias, with a potentially larger impact, may be present in these generators.

## V. A CONTROLLED-BIAS GENERATOR

No generator of minimal puzzles is currently guaranteed to have no bias and building such a generator with reasonable computation times seems out of reach.

We therefore decided to proceed differently: taking the generators (more or less) as they are and applying corrections for the bias, if we can estimate it.

The method was inspired by what is done in cameras: instead of complex optimisations of the lenses to reduce typical anomalies (such as chromatic aberration, purple fringing, barrel or pincushion distortion…) – optimisations that lead to large and expensive lenses –, some camera makers now accept a small amount of these in the lenses and they correct the result in real time with dedicated software before recording the photo.

The main question was then: can we determine the bias of the classical top-down or bottom-up generators? Once again, the answer was negative. But there appears to be a medium way between "improving the lens" and "correcting its small defects by software": we devised a modification of the top-down generators such that it allows a precise mathematical computation of the bias.

### A. *Definition of the controlled-bias generator*

Consider the following, modified top-down generator, the ***controlled-bias generator***; the procedure described below produces *one* minimal puzzle (it has to be iterated n times to produce n minimal puzzles):
```
1) choose randomly a complete grid P;
2a) choose one clue randomly from P and delete it,
set P2 = the resulting puzzle;
2b) if P2 has several solutions, GOTO 1 (i.e.
restart with another complete grid);
2c) if P2 is minimal, printout P2 and exit the whole
procedure;
2d) otherwise (the puzzle has more than one
solution), set P=P2 and GOTO 2a
```

The only difference with the top-down algorithm is in clause 2b: if a multi-solution puzzle is encountered, instead of backtracking to the previous state, the current complete grid is merely discarded and the search for a minimal puzzle is restarted with another complete grid.

Notice that, contrary to the standard bottom-up or top-down generators, which produce one minimal puzzle per complete grid, the controlled-bias generator will generally use several complete grids before it outputs a minimal puzzle. The efficiency question is: how many? Experimentations show that many complete grids (approximately 250,000 in the mean) are necessary before a minimal puzzle is reached. But this question is about the efficiency of the generator, it is not a conceptual problem.

The controlled-bias generator has the same output and will therefore produce minimal puzzles according to the same probability distribution as its following "virtual" counterpart:
```
Repeat until a minimal puzzle has been printed:
1) choose randomly a complete grid P;
2) repeat while P has at least one clue:
   2a) choose one clue randomly from P and delete it,
   thus obtaining a puzzle P2;
   2b) if P2 is minimal, print P2 (but do not exit the
   procedure);
   2c) set P=P2.
```
The only difference with the controlled-bias generator is that, once it has found a minimal or a multi-solution puzzle, instead of exiting, this virtual generator continues along a useless path until it reaches the empty grid.

But this virtual generator is interesting theoretically because it works similarly to the random uniform search defined in the next section and according to the same transition probabilities and it outputs minimal puzzles according to the probability Pr on the set B of minimal puzzles defined below.

*B. Analysis of the controlled-bias generator*

We now build our formal model of this generator.

Let us introduce the notion of a *doubly indexed puzzle*. We consider only (single or multi solution) consistent puzzles P. The double index of a doubly indexed puzzle P has a clear intuitive meaning: the first index is one of its solution grids and the second index is a sequence (notice: not a set, but a sequence, i.e. an ordered set) of clue deletions leading from this solution to P. In a sense, the double index keeps track of the full generation process.

Given a doubly indexed puzzle Q, there is an underlying singly-indexed puzzle: the ordinary puzzle obtained by forgetting the second index of Q, i.e. by remembering the solution grid from which it came and by forgetting the order of the deletions leading from this solution to Q.

Given a doubly indexed puzzle Q, there is also a non indexed puzzle, obtained by forgetting the two indices.

Notice that, for a single solution doubly indexed puzzle, the first index is useless as it can be computed from the puzzle; in this case singly indexed and non indexed are equivalent. In terms of the generator, it could equivalently output minimal puzzles or couples (minimal-puzzle, solution).

Consider now the following layered structure (a forest, in the graph-theoretic sense, i.e. a set of disjoint trees, with branches pointing downwards), the nodes being (single or multi solution) doubly indexed puzzles:

- floor 81 : the N different complete solution grids (considered as puzzles), each indexed by itself and by the empty sequence; notice that all the puzzles at floor 81 have 81 clues;

- recursive step: given floor n+1 (each doubly indexed puzzle of which has n+1 clues and is indexed by a complete grid that solves it and by a sequence of length 81-(n+1)), build floor n as follows:

each doubly indexed puzzle Q at floor n+1 sprouts n+1 branches; for each clue C in Q, there is a branch leading to a doubly indexed puzzle R at floor n: R is obtained from Q by removing clue C; its first index is identical to that of Q and its second index is the (81-n)-element sequence obtained by appending C to the end of the second index of Q; notice that all the doubly indexed puzzles at floor n have n clues and the length of their second index is equal to $1 + (81-(n+1)) = 81-n$.

It is easy to see that, at floor n, each doubly indexed puzzle has an underlying singly indexed puzzle identical to that of $(81 - n)!$ doubly indexed puzzles with the same first index at the same floor (including itself).

This is equivalent to saying that, at any floor $n < 81$, any singly indexed puzzle Q can be reached by exactly $(81 - n)!$ different paths from the top (all of which start necessarily from the complete grid defined as the first index of Q). These paths are the $(81 - n)!$ different ways of deleting one by one its missing 81-n clues from its solution grid.

Notice that this would not be true for non indexed puzzles that have multiple solutions. This is where the first index is useful.

Let N be the number of complete grids (N is known to be close to $6.67 \times 10^{21}$, but this is pointless here). At each floor n, there are $N \cdot 81! / n!$ doubly indexed puzzles and $N \cdot 81! / (81-n)! / n!$ singly indexed puzzles. For each n, there is therefore a uniform probability $P(n) = 1/N \cdot 1/81! \cdot (81-n)! \cdot n!$ that a singly indexed puzzle Q at floor n is reached by a random (uniform) search starting from one of the complete grids. What is important here is the ratio:

$P(n+1) / P(n) = (n + 1) / (81 - n)$.

This formula is valid globally if we start from all the complete grids, as above, but it is also valid for all the single solution puzzles if we start from a single complete grid (just forget N in the proof above). (Notice however that it is not valid if we start from a subgrid instead of a complete grid.)

Now, call B the set of (non indexed) minimal puzzles. On B, all the puzzles are minimal. Any puzzle strictly above B has redundant clues and a single solution. Notice that, for all the puzzles on B and above B, singly indexed and non indexed puzzles are in one-to-one correspondence.

On the set B of minimal puzzles there is a probabily Pr naturally induced by the different Pn's and it is the probability that a minimal puzzle Q is output by our controlled-bias generator. It depends only on the number of clues and it is defined, up to a multiplicative constant k, by $Pr(Q) = k P(n)$, if Q has n clues. k must be chosen so that the probabilities of all the minimal puzzles sum up to 1.

But we need not know k. What is important here is that, by construction of Pr on B (a construction which models the workings of the virtual controlled bias generator), the fundamental relation ***Pr(n+1) / Pr(n) = (n + 1) / (81 - n)*** holds for any two minimal puzzles, with respectively n+1 and n clues.

For n < 41, this relation means that a minimal puzzle with n clues is more likely to be reached from the top than a minimal puzzle with n+1 clues. More precisely, we have:

Pr(40) = Pr(41),
Pr(39) = 42/40 . Pr(40),
Pr(38) = 43/39 . Pr(39).

Repeated application of the formula gives Pr(24) = 61.11 Pr(30) : a puzzle with 24 clues has ~ 61 more chances of being output than a puzzle with 30 clues. This is indeed a strong bias.

A non-biased generator would give the same probability to all the minimal puzzles. The above relation shows that *the controlled bias generator:*
*- is unbiased when restricted (by filtering its output) to n-clue puzzles, for any fixed n,*
*- is biased towards puzzles with fewer clues,*
*- this bias is well known.*
*As we know precisely the bias with respect to uniformity, we can correct it easily by applying correction factors cf(n) to the probabilities on B. Only the relative values of the cf(n) is important: they satisfy cf(n+1) / cf(n) = (81 - n) / (n + 1).* Mathematically, after normalisation, cf is just the relative density of the uniform distribution on B with respect to the probability distribution Pr.

This analysis also shows that a classical top-down generator is still more strongly biased towards puzzles with fewer clues because, instead of discarding the current path when it meets a multi-solution puzzle, it backtracks to the previous floor and tries again to go deeper.

*C. Computing unbiased means and standard deviations using a controlled-bias generator*

In practice, how can one compute unbiased statistics of minimal puzzles based on a (large) sample produced by a controlled-bias generator? Consider any random variable X defined (at least) on minimal puzzles. Define: on(n) = the number of n-clue puzzles in the sample, E(X, n) = the mean value of X for n-clue puzzles in the sample and sd(X, n) = the standard deviation of X for n-clue puzzles in the sample.

The (raw) mean of X is classically estimated as: sum[E(X, n) . on(n)] / sum[on(n)].

The corrected, unbiased mean of X must be estimated as (this is a mere weighted average):
***unbiased-mean(X) =***
  ***sum[E(X, n).on(n).cf(n)] / sum[on(n).cf(n)]***.

Similarly, the raw standard deviation of X is classically estimated as: sqrt{sum[sd(X, n)$^2$ . on(n)] / sum[on(n)]}.

And the unbiased standard deviation of X must be estimated as (this is merely the standard deviation for a weighted average):
***unbiased-sd(X) =***
  ***sqrt{sum[sd(X, n)$^2$.on(n).cf(n)] / sum[on(n).cf(n)]}***.

These formulæ show that the cf(n) sequence needs to be defined only modulo a multiplicative factor.

It is convenient to choose cf(26) = 1. This gives the following sequence of correction factors (in the range n = 19-31, which includes all the puzzles of all the samples we have obtained with all the random generators considered here):

[0.00134  0.00415  0.0120  0.0329  0.0843  0.204  0.464  1  2.037  3.929  7.180  12.445  20.474]

It may be shocking to consider that 30-clue puzzles in a sample must be given a weight 61 times greater than a 24-clue puzzle, but it is a fact. As a result of this strong bias of the controlled-bias generator (strong but known and smaller than the other generators), unbiased statistics for the mean number of clues of minimal puzzles (and any variable correlated with this number) must rely on extremely large samples with sufficiently many 29-clue and 30-clue puzzles.

*D. Implementation, experimentations and optimisations of the controlled-bias generator*

Once this algorithm was defined, it was easily implemented by a simple modification of the top-down suexg – call it ***suexg-cb***. The modified generator, even after some optimisations, is much slower than the original one, but the purpose here is not speed, it is controlled bias.

V. COMPARISON OF RESULTS FOR DIFFERENT GENERATORS

All the results below rely on very large samples. Real values are estimated according to the controlled-bias theory.

*A. Complexity as a function of the generator*

| Generator sample size | bottom-up 1,000,000 | top-down 1,000,000 | ctr-bias 5,926,343 | real |
|---|---|---|---|---|
| SER mean | 3.50 | 3.77 | 4.29 | 4.73 |
| std dev | 2.33 | 2.42 | 2.48 | 2.49 |
| NRCZT mean | **1.80** | **1.94** | **2.22** | **2.45** |
| std dev | 1.24 | 1.29 | 1.35 | 1.39 |

Table 2: SER and NRCZT means and standard deviations for bottom-up, top-down, controlled-bias generators and real estimated values.

| | 0 | 1 | 2 | 3 | 4 | 5 | 6 | 7 | 8 | 9 | 10 | 11 | 12-16 |
|---|---|---|---|---|---|---|---|---|---|---|---|---|---|
| bottom-up | 46.27 | 13.32 | 12.36 | 15.17 | 10.18 | 1.98 | 0.49 | 0.19 | 0.020 | 0.010 | 0 * | 0.01 * | 0 * |
| top-down | 41.76 | 12.06 | 13.84 | 16.86 | 12.29 | 2.42 | 0.55 | 0.15 | 0.047 | 0.013 | 3.8e-3 | 1.5e-3 | 1.1e-3 |
| ctr-bias | 35.08 | 9.82 | 13.05 | 20.03 | 17.37 | 3.56 | 0.79 | 0.21 | 0.055 | 0.015 | 4.4e-3 | 1.2e-3 | 4.3e-4 |
| **real** | **29.17** | **8.44** | **12.61** | **22.26** | **21.39** | **4.67** | **1.07** | **0.29** | **0.072** | **0.020** | **5.5e-3** | **1.5e-3** | **5.4e-4** |

Table 3: The NRCZT-rating distribution (in %) for different kinds of generators, compared to the real distribution.

Table 2 shows that the mean (NRCZT or SER) complexity of minimal puzzles depends strongly on the type of generator used to produce them and that all the generators give rise to mean complexity below the real values.

Table 3 expresses the NRCZT complexity bias of the three kinds of generators. All these distributions have the same two modes, at levels 0 and 3, as the real distribution. But, when one moves from bottom-up to top-down to controlled-bias to real, the mass of the distribution moves progressively to the right. This displacement towards higher complexity occurs mainly at the first nrczt-levels, after which it is only very slight.

*B. Number-of-clues distribution as a function of the generator*

Table 4 partially explains Tables 2 and 3. More precisely, it explains why there is a strong complexity bias in the samples produced by the bottom-up and top-down generators, in spite of the weak correlation coefficient between the number of clues and the (SER or NRCZT) complexity of a puzzle: the bias with respect to the number of clues is very strong in these generators; *controlled-bias, top-down and bottom-up are increasingly biased towards easier puzzles with fewer clues.*

| #clues | bottom-up % | top-down % | ctr-bias % | real % |
|---|---|---|---|---|
| 20 | 0.028 | 0.0044 | 0.0 | 0.0 |
| 21 | 0.856 | 0.24 | 0.0030 | 0.000034 |
| 22 | 8.24 | 3.45 | 0.11 | 0.0034 |
| 23 | 27.67 | 17.25 | 1.87 | 0.149 |
| 24 | 36.38 | 34.23 | 11.85 | 2.28 |
| 25 | 20.59 | 29.78 | 30.59 | 13.42 |
| 26 | 5.45 | 12.21 | 33.82 | 31.94 |
| 27 | 0.72 | 2.53 | 17.01 | 32.74 |
| 28 | 0.054 | 0.27 | 4.17 | 15.48 |
| 29 | 0.0024 | 0.017 | 0.52 | 3.56 |
| 30 | 0 | 0.001 | 0.035 | 0.41 |
| 31 | 0 | 0 | 0.0012 | 0.022 |
| mean | **23.87** | **24.38** | **25.667** | **26.577** |
| std-dev | 1.08 | 1.12 | 1.116 | 1.116 |

Table 4: Number-of-clues distribution (%) for the bottom-up, top-down and controlled-bias generators and real estimated values.

## VI. STABILITY OF THE CLASSIFICATION RESULTS

*A. Insensitivity of the controlled-bias generator to the source of complete grids*

There remains a final question: do the above results depend on the source of complete grids. Until now, we have done as if this was not a problem. But producing unbiased collections of complete grids, necessary in the first step of all the puzzle generators, is all but obvious. It is known that there are $6.67 \times 10^{21}$ complete grids; it is therefore impossible to have a generator scan them all. Up to isomorphisms, there are "only" $5.47 \times 10^9$ complete grids, but this remains a very large number and storing them would require about 500 Gigabytes.

Very recently, a collection of all the (equivalence classes of) complete grids became available in a compressed format (6 Gb); at the same time, a real time decompressor became available. Both were provided by Glenn Fowler. All the results reported above for the controlled bias generator were obtained with this *a priori* unbiased source of complete grids.

Before this, all the generators we tried had a first phase consisting of creating a complete grid and this is where some type of bias could slip in. Nevertheless, several sources of complete grids based on very different generation principles were tested and the classification results remained very stable.

The insensitivity of the controlled-bias generator to the source of complete grids can be understood intuitively: it deletes in the mean two thirds of the initial grid data and any structure that might be present in the complete grids and cause a bias is washed away by the deletion phase.

*B. Insensivity of the classification results to the choice of whips or braids*

In [6], in addition to the notion of a zt-whip, we introduced the apparently much more general notion of a zt-braid, to which a B-NRCZT rating can be associated in the same way as the NRCZT rating was associated to zt-whips. The above statistical results are unchanged when NRCZT is replaced by B-NRCZT. Indeed, in 10,000 puzzles tested, only 20 have different NRCZT and B-NRCZT ratings. The NRCZT rating is thus a good approximation of the (harder to compute) B-NRCZT rating.

## VII. COLLATERAL RESULTS

The number of minimal Sudoku puzzles has been a longstanding open question. We can now provide precise estimates for the mean number of n-clue minimal puzzles per complete grid and for the total number (Table 5).

| number of clues | number of n-clue minimal puzzles per complete grid: mean | number of n-clue minimal puzzles per complete grid: relative error (= 1 std dev) |
|---|---|---|
| 20 | 6.152e+6 | 70.7% |
| 21 | 1.4654e+9 | 7.81% |
| 22 | 1.6208e+12 | 1.23% |
| 23 | 6.8827e+12 | 0.30% |
| 24 | 1.0637e+14 | 0.12% |
| 24 | 6.2495e+14 | 0.074% |
| 26 | 1.4855e+15 | 0.071% |
| 27 | 1.5228e+15 | 0.10% |
| 28 | 7.2063e+14 | 0.20% |
| 29 | 1.6751e+14 | 0.56% |
| 30 | 1.9277e+13 | 2.2% |
| 31 | 1.1240e+12 | 11.6% |
| 32 | 4.7465e+10 | 70.7% |
| **Total** | **4.6655e+15** | **0.065%** |

Table 5: Mean number of n-clue minimal puzzles per complete grid

Another number of interest is the mean proportion of n-clue minimal puzzles among the n-clue subgrids of a complete grids. Its inverse is the mean number of tries one should do to find an n-clue minimal by randomly deleting 81-n clues from a complete grid. It is given by Table 6.

| number of clues | mean number of tries |
|---|---|
| 20 | 7.6306e+11 |
| 21 | 9.3056e+9 |
| 22 | 2.2946e+8 |
| 23 | 1.3861e+7 |
| 24 | 2.1675e+6 |
| 24 | 8.4111e+5 |
| 26 | 7.6216e+5 |
| 27 | 1.5145e+6 |
| 28 | 6.1721e+6 |
| 29 | 4.8527e+7 |
| 30 | 7.3090e+8 |
| 31 | 2.0623e+10 |
| 32 | 7.6306e+11 |

Table 6: Inverse of the proportion of n-clue minimals among n-clue subgrids

One can also get, still *with 0.065% relative error*: after multiplying by the number of complete grids (known to be 6,670,903,752,021,072,936,960 [13]), *the total number of minimal Sudoku puzzles: 3.1055e+37*; after multiplying by the number of non isomorphic complete grids (known to be 5,472,730,538 [14]), *the total number of non isomorphic minimal Sudoku puzzles: 2.5477e+25*.

## VIII. CONCLUSION

The results reported in this paper rely on several months of (3 GHz) CPU time. They show that building unbiased samples of a CSP and obtaining unbiased statistics can be very hard.

Although we presented them, for definiteness, in the specific context of the Sudoku CSP, the sample generation methods described here (bottom-up, top-down and controlled-bias) could be extended to many CSPs. The specific $P(n+1)/P(n)$ formula proven for the controlled-bias generator will not hold in general, but the general approach can in many cases help understand the existence of a very strong bias in the samples. Even in the very structured and apparently simple Sudoku domain, none of this was clear before the present analysis.

## REFERENCES

[1] E.P.K. Tsang, *Foundations of Constraint Satisfaction*, Academic Press, 1993.

[2] H.W. Guesgen & J. Herztberg, *A Perspective of Constraint-Based Reasoning*, Lecture Notes in Artificial Intelligence, Springer, 1992.

[3] D. Berthier: *The Hidden Logic of Sudoku*, Lulu.com, May 2007.

[4] D. Berthier: *The Hidden Logic of Sudoku (Second Edition)*, Lulu.com Publishers, December 2007.

[5] D. Berthier: From Constraints to Resolution Rules; Part I: conceptual framework, *CISSE08/SCSS, International Joint Conferences on Computer, Information, and System Sciences, and Engineering*, December 4-12, 2009.

[6] D. Berthier: From Constraints to Resolution Rules; Part II: chains, braids, confluence and T&E, *CISSE08/SCSS, International Joint Conferences on Computer, Information, and System Sciences, and Engineering*, December 4-12, 2009.

[7] M. Gary & D. Johnson, *Computers and Intractability: A Guide to the Theory of NP-Completeness*, Freeman, 1979.

[10] G. Riley, *CLIPS online documentation*, http://clipsrules.sourceforge.net/OnlineDocs.html, 2008.

[11] E.J. Friedmann-Hill, *JESS Manual*, http://www.jessrules.com/jess/docs/71, 2008.

[12] G. Stertenbrink, *suexg*, http://www.setbb.com/phpb/viewtopic.php?t=206&mforum=sudoku, 2005.

[13] B. Felgenhauer & F. Jarvis, Sudoku enumeration problems, http://www.afjarvis.staff.shef.ac.uk/sudoku/, 2006.

[14] E. Russell & F. Jarvis, There are 5472730538 essentially different Sudoku grids, http://www.afjarvis.staff.shef.ac.uk/sudoku/sudgroup.html, 2006.